\documentclass[11pt]{article}
\usepackage{coling2020}
\usepackage{times}
\usepackage{url}
\usepackage{latexsym}

\usepackage{graphicx}
\usepackage{tabularx}
\usepackage{soul}
\usepackage{booktabs}
\usepackage{multirow}
\usepackage{xcolor}

\usepackage[T1]{fontenc}
\usepackage[utf8]{inputenc}
\usepackage{booktabs}

\usepackage{hyperref}
\usepackage{xstring}
\usepackage{xspace}
\usepackage{covington}

\definecolor{lightblue}{RGB}{200,245,255}
\definecolor{lightred}{RGB}{255,200,200}
\definecolor{lightgreen}{RGB}{205,255,200}
\definecolor{lightgrey}{RGB}{225,225,225}

\newcommand{\mt}{\textsc{Mt}\xspace}
\newcommand{\mtbow}{\textsc{Mt-Bow}\xspace}
\newcommand{\mtbowlex}{\textsc{Mt-Full}\xspace}
\newcommand{\unsup}{\textsc{Unsup}\xspace}
\newcommand{\unsupbow}{\textsc{Unsup-Bow}\xspace}
\newcommand{\unsupbowlex}{\textsc{Unsup-Full}\xspace}
\newcommand{\blse}{\textsc{Blse}\xspace}
\newcommand{\cwe}{\textsc{Cwe}\xspace}
\newcommand{\mono}{\textsc{Mono}\xspace}
\newcommand{\mbert}{\textsc{mBert}\xspace}
\newcommand{\xlm}{\textsc{XLM-Roberta}\xspace}

\newcommand{\ie}{\textit{i.\,e.}\xspace}

\title{Cross-lingual Emotion Intensity Prediction}

\author{Irean Navas Alejo \\
  Expert System \\
  {\tt irean.navas@gmail.com} \\\And
  Toni Badia\\
   Universitat Pompeu Fabra \\
  {\tt tbadia@upf.edu} \\\And
  Jeremy Barnes \\
  University of Oslo\\
 {\tt jeremycb@ifi.uio.no} \\}

\begin{document}
\maketitle
\begin{abstract}

Emotion intensity prediction determines the degree or intensity of an emotion that the author expresses in a text, extending previous categorical approaches to emotion detection. While most previous work on this topic has concentrated on English texts, other languages would also benefit from fine-grained emotion classification, preferably without having to recreate the amount of annotated data available in English in each new language. Consequently, we explore cross-lingual transfer approaches for fine-grained emotion detection in Spanish and Catalan tweets. To this end we annotate a test set of Spanish and Catalan tweets using Best-Worst scaling. We compare six cross-lingual approaches, e.g., machine translation and cross-lingual embeddings, which have varying requirements for parallel data -- from millions of parallel sentences to completely unsupervised. The results show that on this data, methods with low parallel-data requirements perform surprisingly better than methods that use more parallel data, which we explain through an in-depth error analysis. We make the dataset and the code available at \url{https://github.com/jerbarnes/fine-grained_cross-lingual_emotion}.
\end{abstract}

\section{Introduction}


Emotion analysis within natural language processing attempts to identify the private states \cite{Wiebe2005} expressed in written text, which in many cases are only implicitly available. Research often classifies these emotions into discrete categories \cite{Ekman1999-EKMBE,Plutchik2001}, such as \textit{anger}, \textit{fear}, \textit{joy}, or \textit{sadness}. This \emph{discrete} approach to emotion has been applied to fairy tales \cite{Alm2005}, headlines \cite{Strapparava2007}, and more recently micro-blogging services, such as twitter \cite{mommamad-etal-2015,schuff-etal-2017-annotation}. However, people can express emotion in ways that require a more fine-grained approach than the basic discrete version. Take the following two sentences:

\begin{examples}
    \item I am not feeling particularly happy today
    \item I feel like I am the most miserable person on earth
\end{examples}

\blfootnote{
    \hspace{-0.65cm}  
    This work is licensed under a Creative Commons
    Attribution 4.0 International License.
    License details:
    \url{http://creativecommons.org/licenses/by/4.0/}.
}

Both of these examples would be labelled with the emotion \textit{sadness}. However, it is clear that the second sentence expresses a larger degree of sadness than the first, which categorical approaches to emotion analysis would not be able to identify. This motivates the need to move to a more fine-grained approach to emotion analysis. \emph{Emotion intensity prediction} \cite{mohammad-bravo-marquez-2017-wassa} does just this by extending emotion prediction from a classification task to a regression task. Given a text, the goal is to determine a real-valued number between $-1$ and $1$ representing the \emph{intensity} of the emotion present. This approach allows to capture more subtle differences between expressions of emotion.

Current state-of-the-art approaches to emotion intensity are based on supervised machine learning approaches, which combine several sources of annotated corpora, emotion and sentiment lexicons in order to achieve the best performance. However, the combination of all of these necessary resources is only available in a few high-resource languages, with English easily having the largest number. Collecting a similar set of resources for all other languages is prohibitively expensive and would require years of work. Therefore, it would be preferable to find a way to use the available resources in English to perform emotion intensity prediction in other languages.
 
Cross-lingual methods -- either translation or cross-lingual embedding approaches -- offer a possible solution to the lack of labeled data and have shown promise for sentiment analysis at document-level \cite{adan-chen-tacl,chen-etal-2019-multi}, sentence-level \cite{barnes-etal-2018-bilingual,feng-wan-2019-learning}, and fine-grained \cite{hangya-etal-2018-two,Barnes2019EmbeddingPF}. However, the greater number of classes in emotion classification and the difficulty of the regression task means that it is not obvious that the cross-lingual approaches that work well for sentiment analysis will necessarily work for cross-lingual emotion intensity prediction.

In this work, we provide the first attempt at cross-lingual emotion intensity prediction,  by comparing methods which rely on cross-lingual embeddings, machine translation, and unsupervised machine translation to transfer resources from English to predict the emotion in languages that do not have large available datasets or lexicon resources. For testing, we additionally annotate a dataset of tweets in Spanish and Catalan. Our results show that surprisingly unsupervised machine translation is able to outperform both supervised machine translation and cross-lingual embedding methods on these datasets. We additionally perform detailed quantitative and qualitative error analyses of the supervised and unsupervised translation approaches, concluding that while the overall translation quality of the supervised system is better than the unsupervised system, it often does not translate hashtags, which are an important source of emotion information for this task.




In the rest of the paper we discuss related work (Section \ref{sec:related}) and provide a description of the datasets (Section \ref{sec:data}) and models (Section \ref{sec:exp}) used for the experiments. We then discuss the results (Section \ref{sec:results}) and provide an in-depth analysis of why certain models perform better (Section \ref{sec:analysis}).

\section{Related Work}
\label{sec:related}

Emotion detection attempts to identify explicitly or implicitly mentioned emotions in a text, either by following a proposed set of basic emotion categories \cite{Plutchik1980,Ekman1992AnAF,Ekman1999-EKMBE} or through valence-arousal approaches \cite{Russell2003}. However, in contrast to other tasks which also attempt to detect evaluative language, such as subjectivity or sentiment analysis, there are relatively few annotated resources, and most of these resources are found only in English \cite{Alm2005,Aman2007,Strapparava2007,schuff-etal-2017-annotation}. A notable exception is the deISEAR dataset \cite{troiano-etal-2019-crowdsourcing}, which crowdsources descriptions of emotional events in German. 

Annotating categorical emotion is a subjective and complicated task, which often leads to low inter-annotator agreement \cite{schuff-etal-2017-annotation}. However, Best-worst scaling has shown to improve overall agreement scores when annotating tweets \cite{mohammad-bravo-marquez-2017-wassa,mohammad-etal-2018-semeval}. In this approach, annotators are shown $n$ items ($n > 1$, normally 4) and they must choose only the items that are \emph{best} and \emph{worst}, \ie those that most and least represent the phenomena under question. Despite these advances in annotating, for most languages in the world, there exists no annotated emotion dataset which could enable supervised emotion classification.

On English data, previous approaches to classifying emotion have used word and character n-gram features \cite{mohammad-2012-emotional}, sentiment and emotion lexicon features \cite{mohammad2015-hashtags}, as well as a variety of neural networks \cite{koper-etal-2017-ims,felbo-etal-2017-using,bostan-klinger-2019-exploring}. Typically, strong emotion classification systems use a combination of these features to get the strongest performance.

\subsection{Emotion intensity prediction}

Emotion intensity proposes a more fine-grained view of emotion classification. Specifically,  given a tweet and  an  emotion  X, the goal is to determine the intensity or degree of emotion X expressed in the text -- a real-valued score between 0 and 1. This task has already been the topic of two shared tasks \cite{mohammad-bravo-marquez-2017-wassa,mohammad-etal-2018-semeval}, which attracted many participants.

Given the complexity of the task and the relatively small amount of annotated training data available, it is perhaps unsurprising that state-of-the-art methods incorporate information from external sources, either in the form of specialized word embeddings \cite{goel-etal-2017-prayas}, lexicon features \cite{koper-etal-2017-ims,duppada-hiray-2017-seernet}, or transfer learning methods \cite{felbo-etal-2017-using}. 

Additionally, in contrast to related tasks, such as sentiment analysis where end-to-end neural methods often give state-of-the-art results \cite{barnes-etal-2017-assessing,ambartsoumian-popowich-2018-self}, for emotion intensity prediction n-grams, character n-grams, word embedding features, and lexicon features play a more important role \cite{mohammad-bravo-marquez-2017-wassa,koper-etal-2017-ims,duppada-hiray-2017-seernet}. However, it is not clear if the same features are equally important when performing this task crosslingually.



\subsection{Cross-lingual approaches}

For other tasks which classify affective text, such as sentiment analysis, cross-lingual approaches have shown promise for classifying a low-resource target language by leveraging labeled data from high-resource source languages, such as English \cite{barnes-etal-2018-bilingual,chen-etal-2019-multi}.

We divide cross-lingual approaches into machine translation (\mt) techniques and word embedding techniques, as they generally have different data requirements and different models. \mt approaches can either be supervised, which requires parallel corpora, or unsupervised, relying only on monolingual corpora) approaches. While most MT research has focused on resource-rich languages where Neural MT (NMT) has indeed displaced Statistical MT, a recent line of work has managed to train a NMT system without any supervision, relying on monolingual corpora alone \cite{artetxe-etal-2018-unsupervised}. This would be particularly useful for low-resource languages if the translation quality proved good enough to enable a classifier to reliable predict the emotion.

Cross-lingual embedding methods instead require large monolingual corpora, and small amounts of bilingual signal, often only small bilingual lexica. \newcite{barnes-etal-2018-bilingual} uses monolingual embeddings,
bilingual lexicons and jointly learns cross-lingual embeddings
while training an sentiment classifier. The bilingual sentiment
embeddings (\blse) method predicts sentiment of source sentences projecting the vector of the source embeddings into the joint space and repeats the process for the target language using the target embeddings, meaning the original datasets and not the translations, and projecting them into the joint space to obtain the prediction.

More recently, cross-lingual methods have resorted to multilingual language modelling \cite{devlin-etal-2019-bert,conneau-etal-2020-unsupervised} based on pretraining large transformer models \cite{Vaswani2017} on unlabeled text. These models do not explicitly model inter-language representations, but they give surprising cross-lingual performance on many tasks \cite{wu-dredze-2019-beto}.

\section{Datasets}
\label{sec:data}

    
For training the emotion intensity classifiers, we use the English data from the WASSA 2017 shared task on emotion intensity prediction \cite{mohammad-bravo-marquez-2017-wassa}. The authors collected tweets and used crowd-annotation to achieve real-valued labels for four emotions (\textit{anger}, \textit{fear}, \textit{joy}, and \textit{sadness}). We use their predefined splits (statistics are shown in Table \ref{table:statistics}).

\subsection{Annotation of test data}


For testing in the two target languages (Spanish and Catalan) we create two annotated datasets following the methodology of \newcite{mohammad-bravo-marquez-2017-wassa}. We gather tweets (342 Spanish tweets and 280 Catalan tweets) which contained one of the six emotion terms\footnote{We used the following translations of the emotion terms to gather tweets: \textit{felicidad, enfadado, tristeza, asco, miedo, sorpresa} for Spanish and \textit{felicitat, enfadat, tristesa, fàstic, por, sorpresa} in Catalan.} \textit{anger}, \textit{disgust}, \textit{fear}, \textit{joy}, \textit{sadness} and \textit{surprise}). This original download leads to between 385-600 tweets per emotion, totalling 3279 Spanish and 2941 Catalan tweets. These tweets are then filtered and normalized in order to delete the retweets, mentions and links, as well as adverts and tweets that only contained images, resulting in 342 tweets in Spanish and 280 in Catalan.

Following the methodology set out in Best-Worst Scaling (BWS), each annotator is given four items (4-tuple) and is asked which item is the \emph{best} (highest in terms of the property of interest) and which is the \emph{worst} (least in terms of the property of interest)
\cite{kiritchenko-mohammad-2016-capturing}. The annotator must then choose which one of those 4 tweets represents each emotion the most
and which represents it the least. Given the small number of tweets, we include all in the annotation of each emotion. An example of the annotation
process in Spanish is shown in Table \ref{tab:example_fourtuple}.

\begin{table}[]
    \centering
    \begin{tabular}{ll}
    \toprule
         1. & Que lindo fue volver a meterse a nadar hoy  \\
         2. & Hoy estoy triste.... \\ 
         3. & le acabo de romper una pata a la araña y se me subio a
la mano \\
        4. & estoy tan enfadado.... gracias a dios que nadie me
entiende. \\
    \bottomrule
    \end{tabular}
    \caption{An example 4-tuple for the Spanish data for \textit{sadness}.}
    \label{tab:example_fourtuple}
\end{table}

This annotation task presents a number of challenges. For example, annotators cannot simply rely on keywords in the tweet to identify the intensity of an emotion, given that many
times the authors of tweets use emotional hashtags ironically. Additionally, differentiating between four tweets that all have relatively low intensities of an emotion can be difficult.
Finally, inferring the emotion conveyed in a short text is known to be subjective and challenging on its own \cite{schuff-etal-2017-annotation}.

\subsection{Inter-annotator agreement}

\begin{table*}[]
\newcommand{\sep}{\cmidrule(lr){2-3}\cmidrule(lr){4-5}}
    \centering
    \begin{tabular}{lllll}
    \toprule
     & \multicolumn{2}{c}{Catalan} & \multicolumn{2}{c}{Spanish} \\
    \sep
          & Pearson & Spearman & Pearson & Spearman \\
    \sep
    \textit{anger} & 0.68 \tiny{(0.02)} & 0.66 \tiny{(0.03)} & 0.77 \tiny{(0.02)} & 0.78 \tiny{(0.02)}\\
    \textit{disgust} & 0.71 \tiny{(0.02)} & 0.70 \tiny{(0.02)} & 0.76 \tiny{(0.02)} & 0.77 \tiny{(0.02)}\\
    \textit{fear} & 0.69 \tiny{(0.02)} & 0.67 \tiny{(0.02)} & 0.74 \tiny{(0.02)} & 0.74 \tiny{(0.02)}\\
    \textit{joy} & 0.66 \tiny{(0.02)} & 0.64 \tiny{(0.02)} & 0.76 \tiny{(0.02)} & 0.76 \tiny{(0.02)}\\
    \textit{sadness} & 0.65 \tiny{(0.02)} & 0.64 \tiny{(0.02)} & 0.74 \tiny{(0.02)} & 0.75 \tiny{(0.02)}\\
    \textit{surprise} & 0.44 \tiny{(0.03)} & 0.44 \tiny{(0.04)} & 0.65 \tiny{(0.02)} & 0.62 \tiny{(0.02)}\\
    \bottomrule
    \end{tabular}
    \caption{Split-half reliability (as measured by Pearson and Spearman rank correlation) for \textit{anger}, \textit{disgust}, \textit{fear}, \textit{joy}, \textit{sadness}, and \textit{surprise} annotations of tweets in the Tweet Emotion Intensity Dataset.}
    \label{table:splithalf}
\end{table*}

After annotating the tuples, we use split-half correlation to determine inter-annotator agreement (shown in Table \ref{table:splithalf}). We report Pearson correlation, which is a number between -1 and 1 that indicates the extent to which two variables are linearly related, and Spearman correlation, which measures the strength and direction of association between two ranked variables. As shown in Table \ref{table:splithalf}, we obtain strong correlations ($>0.6$) for all emotions except for \textit{surprise}. We find a higher correlation score in Spanish than in Catalan for all emotions. The lower correlation for the Catalan tweets could be due to the fact that there are fewer tweets referring to \textit{joy} and \textit{surprise}, which made the annotation task harder. 
It is important to point out that \textit{surprise} has the lowest scores, which is known to be a difficult emotion to classify \cite{schuff-etal-2017-annotation}, and has even been split into positive and negative surprise \cite{Alm2005}. In fact, we disregard \textit{surprise} and \textit{disgust} for the rest of the experiments, as we have no English annotated data for these emotions. However, the data will be made available with all annotations.

\begin{table*}[]
\newcommand{\sep}{\cmidrule(lr){1-1}\cmidrule(lr){2-3}\cmidrule(lr){4-5}\cmidrule(lr){5-6}}
    \centering
    \begin{tabular}{llllll}
       \toprule
      & EN$_{train}$ & EN$_{dev}$ & EN$_{test}$ & CA$_{test}$ & ES$_{test}$ \\
    \sep
    \textit{anger} & 857 &  84 & 760 & 280 & 342\\
    \textit{fear} & 1147 & 110 & 995 & 280 & 342\\
    \textit{joy} & 823 & 79 & 714 & 280 & 342\\
    \textit{sadness} & 786 & 74 & 673 & 280 & 342 \\
    \bottomrule
    \end{tabular}
    \caption{Statistics of the English train (EN$_{train}$), development (EN$_{dev}$), and Catalan (CA$_{test}$) and Spanish (ES$_{test}$) test sets used in the experiments.}
    \label{table:statistics}
\end{table*}

\section{Experimental Setup}
\label{sec:exp}

In order to determine how much bilingual signal is required to predict crosslingual emotion intensity, we compare four methods with differing data needs. In the following, we describe these methods from those that require the largest amount of bilingual signal (\mt) to those that require the least (\unsup). Each model is trained on EN$_{train}$ and then tested on EN$_{test}$, CA$_{test}$, and ES$_{test}$. EN$_{dev}$ is only used for hyperparameter optimization.


\paragraph{Supervised Machine Translation (\mt):}
We use GoogleTranslate\footnote{Available at \url{https://translate.google.com}.}, which makes use of large amounts of parallel data, to translate the test samples to English. We then train a Support Vector Regression model\footnote{We use the version available in the Sklearn toolkit
\cite{Pedregosa:2011:SML:1953048.2078195}.} on bag of words representations (\mtbow) and a second model with a number of additional features (\mtbowlex). For the \mtbowlex model, we include n-gram features (1-4), character n-grams (3-5), embedding features created by averaging the embeddings of all the tokens in the tweet, and finally features from the following lexicons: NRC Hashtag Sentiment Lexicon \cite{mohammad-etal-2013-nrc}, NRC hashtag emotion association lexicon \cite{mohammad-2012-emotional,mohammad2015-hashtags} and the NRC Word-Emotion Association Lexicon \cite{mohammad-turney2013}, where each feature is a real valued number which represents how much each word is associated to a polarity or emotion. The final representation of each tweet using the \mtbowlex method is therefore a 65860 dimensional vector. Finally, we train the SVR model using the following settings (linear kernel, $C=100$) on the original English training data and test on the translated test set. 


\paragraph{Cross-lingual Word Embeddings (\cwe): }
We create 300 dimensional monolingual word2vec embeddings for source and target languages by training on Wikipedia corpora (see \unsup for more information on the corpora) and then use VecMap \cite{artetxe-etal-2017-learning} to learn an orthogonal projection of the word embeddings to a joint shared embedding space using a small bilingual lexicon\footnote{We use the lexicons made available from \newcite{barnes-etal-2018-bilingual}.} as supervision (5749 and  5310 translation pairs for EN-ES and EN-CA, respectively). Finally, we train a Support Vector Regression model on the source language (EN) using only the crosslingual embeddings as features using the following settings (linear kernel, $C=100$) and test it on the target languages (ES, CA). It is important to highlight that this method does not use the translations but the original texts.

\paragraph{Bilingual Sentiment Embeddings (\blse): }
Like \cwe, this model uses a bilingual lexicon to learn a
mapping from both original vector spaces to a shared bilingual space, but instead jointly learns to predict the sentiment and employs two linear projection matrices. This allows the model to infuse the target embedding space with sentiment information by updating the source space for sentiment and requiring that the target space resemble it as much as possible, using the bilingual dictionary to anchor terms. In this work, we adapt the model to predict emotion intensity by replacing the cross-entropy loss with mean-squared error. We train the model on the English training data and the same bilingual lexicons as for \cwe, optimizing with Adam \cite{Kingma2014a} for 100 epochs with an $\alpha$ of $0.001$. We keep the model with the best performance on the source language development set, and finally test on the target test set. As with \cwe, we highlight that this method does not use the translations but the original texts.

\paragraph{Unsupervised Statistical Machine Translation (\unsup)}

We train an unsupervised statistical machine translation model \cite{artetxe-etal-2018-unsupervised} on Wikipedia corpora for both English-Spanish and English-Catalan. The model first creates monolingual embeddings, then learns to project them to a bilingual space by selecting identical strings as pivots, which serve as a noisy bilingual lexicon, which is improved iteratively. Next, the model induces a noisy phrase table for the SMT model, which is again improved iteratively.

We extract cleaned corpora\footnote{We use the wikiextractor tool available at \url{https://github.com/attardi/wikiextractor}.} from Wikipedia dumps and sentence and word tokenize them, resulting in 89\textasciitilde{}/29\textasciitilde{}/10\textasciitilde{} million sentences for English, Spanish, and Catalan, respectively. We train the \unsup model using the default settings (removing sentences with fewer than 3 and more than 80 tokens, 5-gram language model, 300 dimensional embeddings, 10 rounds of unsupervised tuning for the SMT and 3 rounds of backtranslation). We then translate the test data to English using the \unsup system. Finally, we use bag-of-words representations (\unsupbow) and additional n-gram, character n-gram, embedding, and lexicon information (\unsupbowlex) and train a Support Vector Regression model using the following settings (linear kernel, $C=100$), as we do with the \mt models.

\paragraph{Multi-lingual Language Models}
We use pretrained \mbert and \xlm models to extract features for each example by taking the final \texttt{[CLS]} embedding as the representation for the example. These features are then used to train an SVR model, as with the other experiments. We additionally experimented with adding a linear layer after the final LM layer and fine-tuning the full model, only fine-tuning the linear layer, and using a max pooled representation instead of the \texttt{[CLS]} embedding, but found that these approaches did not perform as well.

\section{Results}
\label{sec:results}
The Pearson correlation results are summarized for all models in Table \ref{table:results} for Spanish and Catalan. We report the individual scores for \textit{anger}, \textit{fear}, \textit{joy}, and \textit{sadness}, as well as the averaged Pearson score of all emotions.

\begin{table*}[]
\newcommand{\sep}{\cmidrule(lr){2-6}\cmidrule(lr){7-11}\cmidrule(lr){12-16}}

\newcommand{\best}[1]{\textbf{#1}}
    \centering
    \resizebox{\textwidth}{!}{%
    \begin{tabular}{lrrrrrrrrrrrrrrr}
    \toprule
        & \multicolumn{5}{c}{Monolingual} & \multicolumn{10}{c}{Cross-lingual} \\
        \cmidrule(lr){2-6}\cmidrule(lr){7-16}
        & \multicolumn{5}{c}{English} & \multicolumn{5}{c}{Catalan} & \multicolumn{5}{c}{Spanish} \\
    \sep
         & \textit{anger} & \textit{fear} & \textit{joy} & \textit{sadness} & avg.  & \textit{anger} & \textit{fear} & \textit{joy} & \textit{sadness} & avg. & \textit{anger} & \textit{fear} & \textit{joy} & \textit{sadness} & avg. \\
    \sep
    \cwe & 0.17 & 0.30 & 0.22& 0.28 & 0.24 & 0.04  & -0.02  & 0.13  & 0.03  & 0.05  & 0.03  & -0.04 & 0.16 & 0.00 & 0.04\\
    \blse & 0.35 & 0.27 & 0.46 & 0.39 & 0.37 & 0.24 & -0.06 & 0.03 & 0.06 & 0.07 & 0.14 & 0.09 & 0.24 & \best{0.12} & 0.15\\
    \mtbow &0.41& 0.51 & 0.48 & 0.47 & 0.47 & 0.28 & 0.10 & 0.19 & 0.02 & 0.15 & 0.14 & 0.06 & 0.17 & -0.11 & 0.07\\
    \unsupbow &0.41& 0.51 & 0.48 & 0.47 & 0.47 & 0.21 & -0.03 & 0.17 & 0.03 & 0.10 & 0.14 & 0.01 & 0.13 & -0.10 & 0.05\\ 
   \mtbowlex & \best{0.60} & 0.60 & \best{0.64} & \best{0.62} & \best{0.62} & 0.37 & 0.35 & \best{0.46} & 0.17 & 0.34 & 0.33 & 0.24 & 0.43 & 0.05 & 0.26 \\
    \unsupbowlex & \best{0.60} & 0.60 & \best{0.64} & \best{0.62} & \best{0.62} & \best{0.42} & 0.40 & 0.45 & 0.21 & \best{0.37} & \best{0.42} & \best{0.31} & \best{0.43} & 0.10 & \best{0.32}\\
    \mbert & 0.20 & 0.27 & 0.33 & 0.31 & 0.27 & -0.05 & 0.06 & 0.03 & -0.15 & -0.03 & 0.12 & 0.13 & 0.05 & 0.0 & 0.08 \\
    \xlm & 0.44 & \best{0.61} & 0.54 & 0.58 & 0.54 & 0.35 & \textbf{0.42} & 0.39 & \best{0.25} & 0.35 & 0.34 & 0.29 & \best{0.43} & \best{0.12} & 0.30 \\
    \bottomrule
    \end{tabular}
    }%
    \caption{ Pearson results of monolingual English-English experiments, as well as cross-lingual English-Catalan and English-Spanish for each emotion and each model. Average column added and best results are shown in \best{bold}.
}
    \label{table:results}
\end{table*}

The approach that obtains the highest overall Pearson
correlation across all emotions on both languages is
\unsupbowlex, averaging 0.37 on Catalan and 0.32 on Spanish. In addition, it is the best performing model on 4 of the 8 tasks, except for Catalan \textit{joy}, where \mtbowlex is 0.01 percentage points (pp.) better, reaching 0.46 and Catalan \textit{sadness} (\xlm is 0.04 pp. better, at 0.25) and  Catalan \textit{fear} (\xlm is 0.02 pp. better, at 0.42) and Spanish \textit{sadness} (\blse is 0.02 pp. better, reaching 0.12). \xlm is the second best model, averaging 0.35 and 0.30 on Catalan and Spanish, respectively, while \mtbowlex is slightly worse (0.34 and 0.26). \mtbow and \unsupbow perform much worse (0.15/0.07 and 0.10/0.05), and the cross-lingual embedding methods are the worst by far (0.04/0.03 for \cwe and 0.07/0.15 for \blse).

It is clear that the additional features (character n-grams, embedding features, and lexicon features) are essential. \mtbowlex performs an average of 0.19 pp. Pearson better than \mtbow, while \unsupbowlex leads to 0.28 pp. improvement over \unsupbow. We further confirm this in Section \ref{sec:ablationstudy}.

Regarding the cross-lingual embedding models, it seems evident that these do not contain enough information to accurately predict emotion intensity in the target language. \blse does outperform \cwe on both Catalan and Spanish (an average of 0.03 pp. and 0.12 pp., respectively) and both \mtbow and \unsupbow on Catalan (0.08 pp. and 0.10 pp.), but the overall performance is still poor. These models are also the poorest performers monolingually.

There is large performance gap between \xlm and \mbert, on both the monolingual (0.27 pp.) and cross-lingual tasks (0.38/0.22 pp.), as \mbert is the weakest cross-lingual model and \xlm the second best.

Additionally, there is a divergence between \textit{sadness} and
the rest of the emotions analyzed, with no model achieving more than 0.21 or 0.12 in Catalan and Spanish. This seems to indicate that sadness may be harder to classify cross-lingually, as monolingually this class is has the best classification results \cite{mohammad-bravo-marquez-2017-wassa,koper-etal-2017-ims}. This class also has the fewest training and development examples in English, which may indicate that the good previous results monolingually may have been due to overfitting to the data. It is also possible that the particulars of the target language test data are the reason for this difference, although the inter-annotator agreement scores suggest that \textit{sadness} is not more difficult than the other classes.

\section{Analysis}
\label{sec:analysis}
In this section we compare both quantitatively and qualitatively the differences in translation quality between \mt and \unsup. Furthermore, we perform an ablation study to determine which features are the most important for \mono, \mtbowlex, and \unsupbowlex.

\subsection{Differences in translation quality}

\begin{table}[]
\newcommand{\sep}{\cmidrule(lr){3-3}\cmidrule(lr){4-4}\cmidrule(lr){5-5}\cmidrule(lr){6-6}\cmidrule(lr){7-7}\cmidrule(lr){8-8}\cmidrule(lr){9-9}\cmidrule(lr){10-10}\cmidrule(lr){11-11}}
    \centering
    \begin{tabular}{llrrrrrrrrrr}
    \toprule
& &hashtags & lexical & insert. & delet. & untrans. & slang & names & nums. & Total \\
\sep
\multirow{2}{*}{CA} & \mt & \textbf{90} & 53 & 2 & 18 &17 &26 &5 &2 & 213\\
                     & \unsup & 60 & \textbf{67} & \textbf{7} & \textbf{14} & \textbf{168} & 29 & \textbf{81} & \textbf{9} & \textbf{435}\\
\sep
\multirow{2}{*}{ES} & \mt & \textbf{62} & 37 & 0 & 4 & 12 & 68 & 0 & 0 & 183\\
                     & \unsup &35 & \textbf{142} & \textbf{13} & \textbf{43} &    \textbf{ 84} & \textbf{101} & \textbf{49} & \textbf{16} & \textbf{467}\\
\bottomrule     
    \end{tabular}
    \caption{Error analysis of the the machine translation used in \mt and \unsup approaches. The error categories include incorrectly translated hashtags, lexical errors, insertions, deletions, untranslated segments, translation errors of slang and non-standard language, mistranslated names and numbers. Number refer to the number of tweets where these errors are found, rather than the number of errors. }
    \label{tab:error_analysis}
\end{table}

Given that twitter is a social network where people express their emotions and opinions on a large variety of topics -- social or personal events, news, and politics -- the translation task is made more difficult. Additionally,  relevant information to emotion classification in tweets is often contained in hashtags
\cite{mohammad-etal-2013-nrc}, which are known to be difficult to translate \cite{gotti-etal-2014-hashtag}. Therefore, cross-lingual approaches to fine-grained emotion detection in twitter are particularly challenging since the language used in twitter usually contains abbreviations, acronyms, emoticons, unusual orthographic elements, slang, and misspellings \cite{liew-turtle-2016-exploring}. All of these phenomena are difficult for both translation- and projection-based cross-lingual approaches.

We manually examine the \mt and \unsup translations of the Catalan and Spanish tweets for translation errors. For each tweet, we determine if there has been an error regarding the hashtags, any lexical errors, insertions, deletions, untranslated segments, errors with non-standard language, errors translating names and errors translating number and show the results in Table \ref{tab:error_analysis}. \mt has fewer errors overall compared to \unsup (213/183 compared to 435/467, respectively) and has fewer of all error types, except for hashtags. For the task of predicting emotion intensity in tweets, the hashtags are often the most informative source, which explains why \unsupbowlex performs better than \mtbowlex in our experiments.

The Spanish translation models generally perform better than the Catalan ones. This is likely due to the larger amount of training data available. However, the Spanish data also contains more use of non-standard language, which is reflected in the \textit{slang} errors. In these cases, \mt generally performs much better than \unsup. 
Interestingly, \unsup tends to mistranslate named entities. Specifically, the model often replaces a named entity with a \emph{similar} entity in the target language. For example, mentions of the Catalan freedom fighter Salvador Puig i Antich are consistently translated to Mumia Abu-Jamal, an American journalist (see Table \ref{tab:example_trans2}). Both were accused of killing a police officer and sentenced to death, which lead to large protests. This is likely due to the nearest neighbor search used to create the original phrase tables.



\begin{table}[]
\newcommand{\names}[1]{{\setlength{\fboxsep}{1pt}\colorbox{lightblue}{#1}}}
\newcommand{\nottrans}[1]{{\setlength{\fboxsep}{1pt}\colorbox{lightred}{#1}}}
    \centering
    \begin{tabular}{ll}
    \toprule
   original & \#\textit{DiosLosCríaYEllosSeJuntan} L'advocat de Camacho en el cas Método 3 \\
            & va redactar la sentència de Puig Antich link  \\
  \mt & \nottrans{\# \textit{DiosLosCríaYesLocated}} The Camacho lawyer in the case Method 3 \\
           & wrote the sentence of Puig Antich link \\
  \unsup &\nottrans{\textit{\# DiosLosCríaYEllosSeJuntan}} \nottrans{l 'advocat} of camacho in the case história 3  \\
       & drafted the verdict of \names{abu-jamal} link \\
  manual trans. & \#\textit{BirdsOfAFeatherFlockTogether} Camacho's lawyer in the Método 3 case \\
  & is the one who sentenced Puig Antich link \\
    \bottomrule
    \end{tabular}
    \caption{An example of a tweet in Catalan (original), its translations using the two machine translation systems (\mt, \unsup), as well as a manual translation. \nottrans{\textit{Untranslated tokens}} are highlighted in red, while \names{\textit{entity errors}} are highlighted in blue.}
    \label{tab:example_trans2}
\end{table}

\begin{table}[]
\newcommand{\hashtag}[1]{{\setlength{\fboxsep}{1pt}\colorbox{lightgrey}{#1}}}
\newcommand{\lexical}[1]{{\setlength{\fboxsep}{1pt}\colorbox{lightgreen}{#1}}}
    \centering
    \begin{tabular}{ll}
    \toprule
         original &  Harto de la situación en \#\textit{Cataluña}. Votamos
mayoritariamente a delincuentes \\ & y tenemos lo que merecemos. No hay solución \#\textit{verguenza} \\
         \mt & Fed up \lexical{of} the situation in \# \textit{Catalonia} . We vote
mostly criminals and we have \\ & what we deserve. There is no solution \hashtag{\textit{\# verguenzak}} \\
         \unsup &  Fed up with the situation in \# \textit{catalonia} . \lexical{They} voted overwhelmingly to criminals \\ & and we have what \lexical{they} deserve no solution \# disgrace   \\
manual trans. & Tired of the situation in \#\textit{Catalonia}. We mainly vote for criminals \\
   & and get what we deserve. There's no solution. \#\textit{shame} \\
    \bottomrule
    \end{tabular}
    \caption{An example of tweet in Spanish (original) and its translation using the two machine translations systems (\mt, \unsup). \hashtag{\textit{Hashtag translation errors}} are highlighted in grey and \lexical{\textit{lexical errors}} are highlighted in green.}
    \label{tab:example_trans3}
\end{table}


Besides the mistranlation of named entities, in Table \ref{tab:example_trans2} we can also see that the multiword hashtag, which contains information necessary to properly interpret the emotional content of the tweet, has not been translated by \mt or \unsup. Note that although this problem could be improved by properly segmenting the hashtags in a previous step \cite{declerck-lendvai-2015-processing,celebi-ozgur-2016-segmenting}, translation would still likely lead to a loss of information \cite{gotti-etal-2014-hashtag} important for emotion classification.

Table 
\ref{tab:example_trans3}, instead,
shows an example from the Spanish dataset where, even though the \mt version better preserves the semantics of the original tweet, it did not correctly translate the emotional hashtag, while \unsup did. On the other hand, for cases where \mtbowlex has better performance, translation quality tends to be the main factor. Specifically, \unsupbowlex tends to leave many words untranslated.

\subsection{Ablation study}
\label{sec:ablationstudy}
In order to determine which features are most predictive for emotion intensity, we perform an ablation study of \mono, \mtbowlex, and \unsupbowlex on the Catalan test data\footnote{The ablation study results for the Spanish data are similar.}. Specifically, we remove a single feature at at time, except for -all lex, where all lexicon features are removed. We include \mono as an upper-bound to determine what features are most important for the task, given enough monolingual data, but note that the test data is different for \mtbowlex and \unsupbowlex, so the exact results are therefore not strictly comparable. The results are shown in Table \ref{table:ablationstudy}.

In general, the cross-lingual models exhibit the same relationship to the features as the monolingual model, although with generally lower performance. Token n-grams are the most important feature for \textit{anger} and \textit{fear}, although less important for \textit{joy} and \textit{sadness}. Word embedding features seem to contribute nothing to the performance. Character n-gram features, on the other hand, contribute little, or even hurt the performance (removing them actually improves the results for \mtbowlex and \unsupbowlex on \textit{fear}, and for \mono and \unsupbowlex on \textit{sadness}). Finally, the lexicon features are important for all emotions.

For \textit{joy}, removing any one set of features does not lead to large drops in performance. Given the good performance of all models on this emotion, it seems to indicate that this class is the easiest to predict, and that the features are relatively redundant. However for \unsupbowlex, removing all lexicon features still leads to a drop of 0.16 pp.

\textit{Sadness} is the most difficult emotion, with the cross-lingual models performing on par with \mono. The lexicon features are the most informative for all models, specifically the NRC sentiment lexicon features (-sent). This effect is even stronger cross-lingually, where without them, the models perform at chance level.

\begin{table*}[]
\newcommand{\sep}{\cmidrule(lr){3-3}\cmidrule(lr){4-4}\cmidrule(lr){5-5}\cmidrule(lr){6-6}\cmidrule(lr){7-7}\cmidrule(lr){8-8}\cmidrule(lr){9-9}\cmidrule(lr){10-10}}
    \centering
    \begin{tabular}{lrrrrrrrrr}
    \toprule
   & & ALL & -ngrams & -char & -embs & -hashtag & -emo & -sent & -all lex\\
    \sep
\multirow{3}{*}{\textit{anger}} & \mono & 0.60 &  \textbf{-0.31} & -0.04 & -0.01 & \underline{-0.06} & -0.01 & -0.01 & \underline{-0.06}\\
                       & \mtbowlex     & 0.37 & \textbf{-0.30} & -0.00 & -0.00 & -0.02 & -0.00 & -0.02 & \underline{-0.07}\\
                       & \unsupbowlex   & 0.42 & \textbf{-0.23} & -0.05 & -0.00 & -0.02 & -0.01 & -0.05 & \underline{-0.11} \\
    \sep
\multirow{3}{*}{\textit{fear}} & \mono  & 0.60 & \textbf{-0.30} & -0.01 & -0.00 & -0.01 & -0.00 & -0.01 & \underline{-0.04}\\
                       & \mtbowlex     & 0.35 & \textbf{-0.31} & +0.06 & -0.00 & -0.02 & -0.01 & -0.05 & \underline{-0.14} \\
                       & \unsupbowlex   & 0.40 & \textbf{-0.26} & +0.03 & -0.00 & -0.01 & -0.01 & -0.08 & \underline{-0.21} \\
    \sep
\multirow{3}{*}{\textit{joy}} & \mono          & 0.64 & \underline{-0.03} & \underline{-0.03} & -0.00 & -0.00 & -0.02 & \underline{-0.03} & \textbf{-0.06} \\
                       & \mtbowlex    & 0.46 & \underline{-0.07} & -0.02 & -0.00 & -0.01 & -0.02 & -0.02 & \textbf{-0.11} \\
                       & \unsupbowlex & 0.45 & \underline{-0.05} & -0.03 & -0.00 & -0.01 & -0.02 & -0.04 & \textbf{-0.16} \\
    \sep
\multirow{3}{*}{\textit{sadness}} & \mono      & 0.62 & -0.01 & \underline{-0.02} & -0.00 & -0.00 & -0.00 & -0.01 & \textbf{-0.04} \\
                       & \mtbowlex    & 0.17 & -0.01 & -0.01 & -0.00 & -0.01 & -0.00 & \underline{-0.09} & \textbf{-0.15}\\
                       & \unsupbowlex & 0.21 & -0.02 & +0.05 & -0.00 & -0.02 & -0.02 & \underline{-0.06} & \textbf{-0.20} \\
    \bottomrule
    \end{tabular}
    \caption{Ablation study of \mono (used to show an informative monolingual baseline), \mtbowlex and \unsupbowlex on the Catalan dataset, where we show the drop in performance (Pearson correlation) when we remove only a single feature at a time (except for -all lex, where all lexicon features are removed). We show the largest drop in \textbf{bold} and the second largest \underline{underlined}. On most emotions (\textit{anger}, \textit{fear} \textit{joy}), removing the n-gram feature leads to the largest drop both mono- and cross-lingually. For \textit{sadness}, however, the NRC sentiment lexicon features (sent) are most decisive.}
    \label{table:ablationstudy}
\end{table*}

\section{Conclusion}


In this paper, we provided the first attempt at cross-lingual emotion intensity prediction, by comparing methods which rely on differing amounts of cross-lingual signal, ranging from millions of parallel sentences (\mt), small bilingual dictionaries (cross-lingual embeddings), to no explicit cross-lingual signal at all (\unsup). We compare these methods on two target languages, Spanish and Catalan, which do not have large available emotion datasets or lexicon resources. In order to test the models, we additionally annotated a small dataset of tweets in Spanish and Catalan. 

Our results show that translation methods outperform embedding-based methods for almost all emotions and achieve reasonable average results, although there is still a noticeable gap to reach monolingual levels. Surprisingly, unsupervised translation is the best performing cross-lingual method, largely due to the fact that it more accurately translates hashtags. \xlm performs nearly as well, but unfortunately cannot be combined with sentiment and emotion lexicons available in English, as it processes the original target-language data. These results may not hold for other domains, such as literature or opinion pieces, where emotional information is not concentrated in a similar way. 

In the future, it would be interesting to perform experiments on various domains, in order to determine whether unsupervised machine translation for cross-lingual emotion is robust to domain shift. Theoretically, this is simpler for unsupervised \mt rather than supervised \mt, which could motivate further research in this direction. As lexicon information has proven so useful for this task, it could be interesting to look into approaches that use this information to improve pretrained multilingual language models. Additionally, given the importance of hashtags for emotion detection in tweets, it would be important for future work in cross-lingual emotion detection to concentrate on achieving better translations of hashtags.

Finally, we contemplate promising research on emotion detection and classification using the newly annotated data in Catalan and Spanish we introduce here. We expect that this will contribute to furthering research on these two languages.



%


\bibliographystyle{acl}
\bibliography{lit.bib}


\end{document}